\def\year{2019}\relax
\documentclass[letterpaper]{article}
\usepackage{aaai19}
\usepackage{times}
\usepackage{helvet}
\usepackage{courier}
\usepackage{graphicx}
\usepackage{color}
\usepackage{tikz}
\usetikzlibrary{mindmap,trees}
\usepackage{smartdiagram}
\usepackage{hyperref}
\usepackage{soul}

\usesmartdiagramlibrary{additions} 
\usetikzlibrary{arrows} 

\long\def\omitit#1{}

\begin{document}
%
\title{Automated Text Summarization for the Enhancement of Public Services }
\author{Xingbang Liu, Janyl Jumadinova\\
Allegheny College \\
Department of Computer Science \\
Meadville, PA 16335 \\
}
\maketitle
\begin{abstract}
Natural language processing and machine learning algorithms have been shown to be effective in a variety of applications. In this work, we contribute to the area of AI adoption in the public sector. We present an automated system that was used to process textual information, generate important keywords, and automatically summarize key elements of the Meadville community statements. We also describe the process of collaboration with  My Meadville administrators during the development of our system. My Meadville, a community initiative, supported by the city of Meadville conducted a large number of interviews with the residents of Meadville during the community events and transcribed these interviews into textual data files. Their goal was to uncover the issues of importance to the Meadville residents in an attempt to enhance public services. Our AI system cleans and pre-processes the interview data, then using machine learning algorithms it finds important keywords and key excerpts from each interview. It also provides searching functionality to find  excerpts from relevant interviews based on specific keywords. Our automated system allowed the city to save over 300 hours of human labor that would have taken to read all interviews and highlight important points. Our findings are being used by My Meadville initiative to locate important information from the collected data set for ongoing community enhancement projects, to highlight relevant community assets, and to assist in identifying the steps to be taken based on the concerns and areas of improvement identified by the community members.

\end{abstract}

\vspace{-0.1in}	
\section{Introduction}

Following the successful implementation of Artificial Intelligence (AI) technologies in the private sector companies, the government agencies have started to adopt AI techniques for different applications such as health care \cite{sun2019mapping}, public safety \cite{kouziokas2017application}, social welfare \cite{Capgemini2017}, and education \cite{timms2016letting}. These AI applications have benefits of cost savings, increase of public employees' productivity by reduction of their workload, new employment opportunities, solutions to the resource allocation problems and enhancement of citizens' satisfaction, but they also present challenges in their successful implementation and use \cite{wirtz2019artificial}. 
AI in the public sector is still a young, emerging field of research and continued extensive research is needed to fully explore the full potential of AI in the public sector, and leverage
various AI technologies to address important problems/needs. A report by Harvard \cite{mehr2017artificial} specifies six types of government problems that could benefit from AI applications the most: resource allocation, large data sets, experts shortage, predictable scenarios, procedural and repetitive tasks, diverse data aggregation and summarization.  This paper makes a contribution in this direction: we present an intelligent system for knowledge extraction from a large textual data set to address the important issue of public service enhancement. We also demonstrate a successful application of this AI technology  in the  city of Meadville and discuss the process of engaging  various stakeholders in the city government and  the challenges that we have encountered during this process.

The city of Meadville  located in 
in Northwestern Pennsylvania has faced many of the same
issues challenging small rural towns across our country, such as population decline,
a dwindling tax base and economic turbulence. In order to investigate and combat some of these issues, in 2015 the City of Meadville applied for and received a grant from the Pennsylvania Humanities Council and the Orton Family Foundation.  The goals of the grant were to build a greater connection with Meadville residents, develop a stronger sense of community identity, and a vision for the future rooted in what matters most to community members. My Meadville Heart and Soul initiative was formed to accomplish the aims identified in this work.
Hosted by the Redevelopment Authority of the City of Meadville, the My Meadville Heart and Soul initiative   interviewed over 700 Meadville residents to learn which places, issues and services residents care about the most. The recorded interviews were then transcribed and collected into a single data set. 

In this project, we first cleaned the data procured by  My Meadville initiative and then used it in our knowledge extraction system. Our goal was to develop a system that could be used by the city of Meadville and relevant community members and organizations to get insight and context into the good and important work that is being done in Meadville, and uncover ways to improve  public relations and services that are meaningful to Meadville residents.  Using natural language processing techniques, such as sentiment analysis and named entity recognizer, and machine learning algorithms, our system  processes  community statements provided by My Meadville, finds important keywords, and then produces a summary of the key excerpts from the data. With the support of the City of Meadville and City Council, our findings are used as an aid by My Meadville initiative to develop community value statements, highlight relevant community assets and to develop an action plan based on the concerns and areas of improvement identified by the community members. An example of the value statement is as follows:

\begin{quote}
\textbf{Children and Youth}: We value youth-centered programming and safe, accessible spaces that support youth and prepare them for a fulfilling future. \\

\textbf{Supporting Data}: \\
\emph{Youth-Centered Programming}: \\
``... Value diverse accessible opportunities for children to engage in and build community.'' \\
``... Appreciates a community that connects resources to education to families to concerns about the whole child through multiple systems ... '' \\
``... Cares about options for youth activities that keep youth occupied, such as a skating rink or pool hall ...'' \\
``... Appreciates teen-friendly activities that are affordable, diverse, and purposeful ...''

\emph{Safe and accessible spaces}: \\
``... Values playgrounds and other safe places for kids to play ...'' \\
`... Treasures a community that offers a variety of safe inside and outside recreational activities for the children and youth of Meadville.'' \\
``... Appreciates activities out of school for students to do so they do not get involved with criminal activities.'' \\
``... Appreciates safe community events, places, and activities for families that bring people together ...'' \\
``... Appreciates walkable options for youth engagement so that families don't stress over transportation ...''

\emph{Support Youth}: \\
``... Values a community that appreciates the voices and input of our youth and nurtures their new ideas ...'' \\
``... Appreciates the local opportunities for youth engagement and the ways in which they are connected to each other ...'' \\
``... Appreciates a community that values the voices of children and youth and provides avenues for them to show their giftedness and to express themselves in meaningful and beneficial ways ...''
\end{quote}

Some examples of the \textbf{Action Plan} items related to the action plan above include:
\begin{quote}
	\begin{enumerate}
		\item  Create a group dedicated to creating an empowered network of teens within the community. 
 		\item Sustain the summer parks program.
 		\item Create a map of public parks.
 		\item Increase available transportation for youth.
	\end{enumerate}
\end{quote}

We also developed a searching functionality of our automatically generated summary of the community statements, where  various organizations and city governmental agencies are able to search the summaries for specific information. We then trained My Meadville administrator to use our system with this searching functionality, which automatically generates a Markdown file on the web for  easy access and interpretation of the search results. In 2019, this extended functionality was utilized to gain insight into the previous summer parks program in Meadville from the residents' stories and to use that information to revitalize and enhance this program in Meadville. Additionally, a separate research study into the museum creation  used our system to uncover the need and desire for such a project in Meadville.

To summarize, the contributions of our work presented in this article are as follows:
\begin{itemize}
	\item An automated system that can process textual information, generate important keywords, and automatically summarize key findings of the large amounts of text. 
	\item An additional search tool that allows users to search the text for specific information and generates results in an easily readable format.
	\item A use case of our system by  My Meadville initiative to aid in  generating value statements and action plans and by the city administrators to study the development of specific projects in the city of Meadville.  
	\item A description of the process of collaboration with the public officials and My Meadville volunteers to ensure transparency and to build trust.
\end{itemize}

\vspace{-0.1in}	
\section{Related Work}
AI in the public sector is a relatively young field but there have been a number of articles demonstrating its use and outlining potential challenges. For example, in \cite{androutsopoulou2019transforming}, the authors present a new approach in the use of chatbots in the public sector to improve the communication between the government and citizens. The presented approach is  built using natural language processing, machine learning and data mining technologies and it  develops a digital channel of communication between the government and citizens. This digital channel uses existing data collected from documents containing legislation and directives, structured data from government agencies' operational systems, and social media data to  facilitate and promote information seeking and conducting of transactions.  The presented approach was validated through a series of application cases with the cooperation of the three Greek government agencies. Given a number of interesting research articles such as the one described above, which demonstrate successful deployment of AI in the public sector, 
Wirtz \emph{ et al.} \cite{wirtz2019artificial}  analyzed and summarized  scientific literature related to AI applications in the public sector. They categorize the research of AI in the public sector as follows: (1) AI government service, (2) working and social environment influenced by AI, (3) public order and law related
to AI, (4) AI ethics, and (5) AI government policy. They also  identify specific AI applications that are valuable for the public sector and present a Four-AI-Challenges Model that incorporates main aspects of AI challenges. Valle-Cruz  \cite{valle2019review} investigate the AI trends in the public sector by surveying 78 recent papers related to this area. Their findings indicate that only normative and exploratory research articles have been published so far and that many public policy challenges face this research area. The authors, however, also outline various benefits of AI application  in public health,  policies on climate change, public management and decision-making, improving government-citizen interaction, personalization of services, analyzing large amounts of data, detecting abnormalities and patterns, and discovering new solutions through modeling and simulations.

There have also been a number of articles proposing some  theoretical frameworks for the successful adoption of AI in the public sector. Chen \emph{et al.} \cite{chen2019ai} presents a four-stage model for AI development in the public sector in order to guide public administrators in its use and navigate the impact AI would have on their organizations. The authors present an application case of AI for delivering public services in local government in China. The outcomes of their  application of AI in  the local government in China could present the research community with the  longitudinal study for  AI  in the public sector. In \cite{engin2019algorithmic}, the authors present a taxonomy of government services to provide an overview of data science automation being deployed by governments world-wide. They present a review of the studies and projects across the world and propose a technological framework on the development of the AI technologies to transform the public sector. 
Our work belongs to the category of research articles that describe the successful development and deployment of AI technology for the public sector use. Our developed AI technology builds upon several existing natural language processing techniques, such as lemmatization and the named entity recognition, uses a number of machine learning algorithms for sentiment analysis and word, phrase and sentence extraction, and builds upon existing open-source projects, such as PyTextRank \cite{PyTextRank}. 

Automatic keyword and keyphrase extraction is the process of selecting words and phrases from the text that can  project the core sentiment of the whole text automatically, and it has become one of the fundamental steps in information retrieval, text mining, and natural language processing applications \cite{siddiqi2015keyword}. In \cite{beliga2015overview}, the authors present the survey for the task of the keyword extraction, concentrating on the graph-based methods. Graph-based representation of text allows for the document to be modeled as a graph, where words are represented as nodes and their relations are represented as edges. They find that graph-based keyword extraction techniques are domain and language independent, thus making them robust and easy to apply to knowledge extraction problems, such as text classification, search, and summarization. 

Text summarization  is a process of extracting the most important features of a text and compiling it into a short text \cite{eduard1998automated}. Various approaches to text summarization have been proposed \cite{aggarwal2012mining}. In a  query-based text summarization a specific portion of the text is utilized to extract the essential keyword from input document to make the summary of corresponding document.
Extractive text summarization process identifies  important information (words or sentences) from the input text using statistical and linguistic features of the sentences and makes the summary of the corresponding document using the most relevant sentences \cite{gupta2010survey}. Similar to this technique, abstractive text summarization finds the important sentences in the text but it uses new concepts and expressions to describe it  by generating a new shorter text that conveys the most important information from the original text document. 

Supervised and unsupervised machine learning techniques have been used for the text summarization task. 
In supervised learning based text summarization labeled data sets are used for training. For example, in \cite{thomas2016automatic}, the authors used a hidden Markov model to automatically extract keywords for text summarization and used human annotated keyword data set to train their model.
However, it is often difficult to find enough labeled data for training, thus, Wong \emph{et al.}  \cite{wong2008extractive}  developed a semi-supervised learning method for co-training
by combining labeled and unlabeled data. They demonstrate that their method is comparable to the supervised learning approach but it only requires about half of the labeling time cost. 
Unsupervised learning based text summarization does not require any labeled data to be used in training. For example, in \cite{garcia2008text}, the authors used a k-means clustering algorithm to generate groups of similar sentences and the most representative sentence was further selected for the summary. 

In our work, we use graph-based keyword, keyphrase and sentence extraction technique and use an unsupervised learning based method for text summarization of  My Meadville community statements provided in the interview data.

\vspace{-0.1in}	
\section{AI System for  My Meadville Interviews}

\begin{figure*}
\begin{center}
	\includegraphics[scale=0.8]{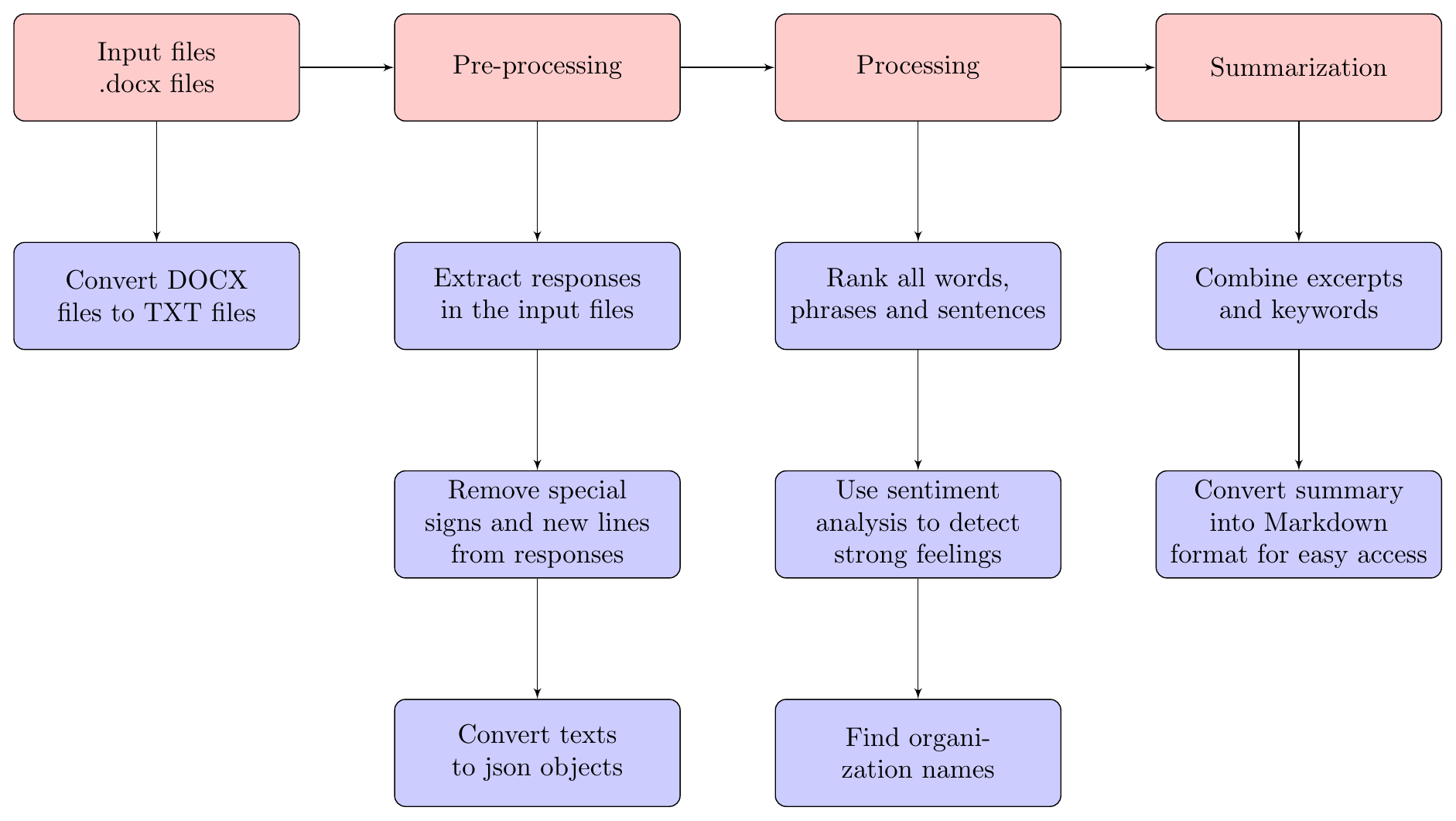}
\end{center}
\caption{Different stages of our system}
\label{fig:flow}
\end{figure*}

\vspace{-0.1in}	
\subsection{Development Process}
After the data was collected by My Meadville we were approached to find ways of automating the process of knowledge extraction. With over 700 text documents, the scarcity of trained volunteers and the lack of funding for professional help, they were seeking ways to make the knowledge extraction from this data feasible. In our initial discussions we identified the goals that the City of Meadville had and possible technological solutions for the data knowledge extraction. We decided to follow an agile development method, with a feedback loop for  My Meadville administrators at the design stage and various development stages. Over a period of a year in 2018, we continued to meet with My Meadville initiative administrators and adjust our development given their feedback. Once our system was fully implemented and tested we gave a demonstration to the My Meadville committee. We then discussed various avenues for adoption and extended use of our system. Following these discussions we prepared and conducted training for the My Meadville committee and its volunteers on using our system and interpreting its results.

\vspace{-0.1in}	
\subsection{My Meadville Interview Data}

In 2015, My Meadville initiative committee members recruited and trained dozens of volunteers to perform interviews and to transcribe them. Then, for a period of two years the volunteers conducted interviews with hundreds of  Meadville residents at various community events, local schools, businesses and organizations. These audio interviews were manually transcribed into Word text files by the volunteers and compiled into a data set that was then used by our system.

Upon reviewing the data set we found that the interview transcriptions did not conform to a single format. For example, in some transcriptions the responses were identified as such, and in others they were not, and we needed to read the document to decide which text was the interviewer comments and questions and which belongs to the interviewee. Since the goal of My Meadville was to extract information from the interviewees, our system only uses the responses in the interviews. Therefore, before using the collected data set in our system we manually looked through each document and whenever necessary made appropriate tags to identify the interviewee response so that it could be recognized by our system. 

All participants whose voices are included in the final data set have signed a waiver that allows My Meadville to use their stories to continue to strengthen the values and visions of our community. Also, all data is anonymous, where all identifying information in the interviews was removed. The hope of My Meadville was that the collected information can provide insight and context into the good and meaningful work that is being done in Meadville, and that this data can be used as a tool to improve the efficiency, purpose, and relationships that make such work possible.

\vspace{-0.1in}	
\subsection{Intelligent Text Extraction and Summarization}

Our AI system processes the textual My Meadville community statements, finds important keywords, and then produces a summary of the key excerpts from all data. It utilizes several open source projects, such as Stanford Named Entity Recognizer (NER), Scikit-learn Sentiment Analysis, and PyTextRank, and  consists of multiple stages as identified in Figure \ref{fig:flow}. The text documents provided by My Meadville were saved as .docx files. We first automatically convert them to .txt files \cite{extractdocx}. Then we go through a pre-processing stage, where the responses are extracted from the interviews, and then cleaned by removing special characters, signs, etc. Finally, the pre-processed interview responses are  converted to JSON files for use in the subsequent stages. 

During the processing stage the text is analyzed and ranking and sentiment analysis are conducted. The text rank layer builds a word graph for voting on the importance of the word based on the baseline approach outlined in \cite{mihalcea2004textrank}. This graph model has the ability to extract key phrases and rank the phrases and sentences. Ranking is done through the idea of voting, where a vote is a connection of one node  to another, that is when one node links to another node it is essentially voting for that node. The higher the number of votes for a node, the higher the importance
of the node. The model also takes the account  the information on the importance of the vote itself, with the TextRank score of a node being calculated based on the votes it receives and the score of the nodes voting for that node. 

\begin{center}
	\begin{figure}
	\large{\textbf{Layer 1}}: \\
		\fcolorbox{black}{gray}{\includegraphics[scale = 0.25]{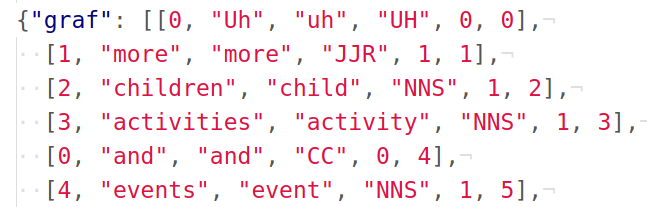}} \\
		
	\large{\textbf{Layer 2}}: \\
         \fcolorbox{black}{gray}{\includegraphics[scale = 0.25]{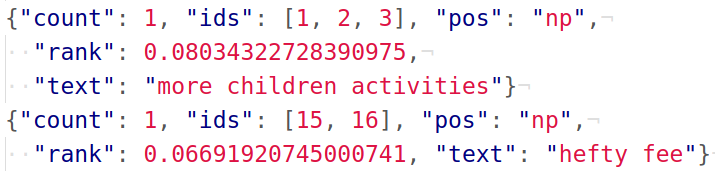}} \\
         
	\large{\textbf{Layer 3}}: \\
     \fcolorbox{black}{gray}{\includegraphics[scale = 0.25]{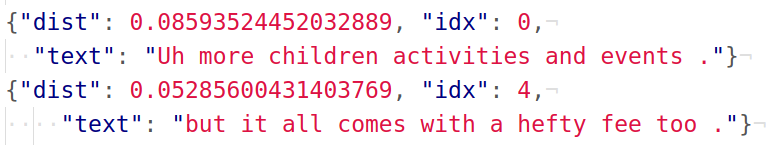}}
         \caption{Example: TextRank layers}
         \label{fig:layers}
	\end{figure}
\end{center}

We build our system based on the implementation of the TextRank method  \cite{PyTextRank} with modifications. This implementation is written in Python and uses spaCy \cite{spacy2}, NetworkX \cite{team2014networkx}, datasketch \cite{datasketch} tools.  The importance calculation is performed in three layers. In the first layer,  statistical parsing and tagging is performed on a document in a JSON format. An example output of this layer can be seen from the top image in Figure \ref{fig:layers}. The second layer collects and normalizes the key phrases from the document produced by the previous step. Finally, during the third layer  a score for each sentence is calculated using  the Jaccard distance between key phrases determined by TextRank and each of its sentences.  The middle and bottom images in Figure \ref{fig:layers} show example outputs of the second and third layers.

The feature vectors built by the text ranking method produced ranked key phrases and sentences as seen in Figure  \ref{fig:layers}. This output  was combined with the output from the sentiment analysis to determine top ranked sentences to be used in extractive summarization of the document.  Scikit-learn machine learning library \cite{Scikit-learn_Sentimental_Analysis} was used to perform sentiment analysis and  to produce a sentiment score. First, training on the data  from Twitter and UCI machine learning database was done with the  use of the pickle module to store the trained model. The UCI data contains sentiment labeled sentences  from Amazon, Yelp, and IMDB contained in {\tt positive.txt} and {\tt negative.txt} files. In order to split the data into training and testing sets, we calculated the frequency distribution of each word for both positive and negative sentences. The top 5,000 words were kept as features, then the features were pickled into one file and shuffled. The testing data included the last 10,000 features, whereas the training data included the previous 10,000 features. Once the training was performed, six algorithms were used for testing, including naive Bayes, multinomial naive Bayes, Bernoulli naive Bayes, logistic regression, linear support vector clustering, and stochastic gradient descent classifier. These algorithms were used to predict the sentiment of the text. In this work we considered text important (high rank) if the outcome of sentiment learning indicated that the text has strong feelings (high score). The sentiment score was represented as a floating-point number and ranged from 0 (negative) to 1 (positive), where having a score closer to 0 or 1 indicated strong feelings.  For each classifier we obtained the accuracy score from {\tt nltk} and the results from the best algorithm were chosen, where the key phrases and sentences with strong positive or negative feelings were selected to be included in the summary.

We also leveraged the Named Entity Recognition technique before producing the summarization of the document. Stanford Named Entity Recognizer (NER) \cite{finkel_grenager_manning_2005} uses an advanced statistical learning algorithm to extract named entities. The NER has three classes, including person, organization, and location entities. In our application, upon consultation with the My Meadville administrators, only organization entities were detected.

After text ranking, sentiment analysis and named entity recognition, all key phrases and sentences determined to be important by  these three techniques are extracted for summarization. 

 
\vspace{-0.1in}	
\section{Experimental Results}
\vspace{-0.05in}	

\begin{table}[ht!]
	\begin{center}
     	\begin{tabular}{|l|l|}
     	\hline 
                    Data set & 706 interviews \\
                    Average interview & 9,161 words \\
                    Shortest interview  & 334 words \\
                    Longest interview & 30,465 words\\
          \hline 
          \end{tabular}
          \caption{Information about the data set used in the experiments}
          \label{tbl:data}
	\end{center}
\end{table}
\vspace{-0.05in}	

The implementation of our system was written in Python 3.6.7 and the experiments were run on Ubuntu Linux 4.15.0. The overview of the data set used in our experiments is shown in Table \ref{tbl:data}. To evaluate the correctness of our system, we have first tested it on five text documents randomly selected from our data set. We carefully read and annotated the text and then manually compared the output produced by our system with our annotations to verify its accuracy.

After manual testing, we ran our system on a complete data set, which produces a textual summary and a list of keywords for each interview document. Two examples of the output produced by our system are shown in Figure \ref{fig:output}, where extracted summary and a list of keywords are included for each of the two documents. The top output corresponds to the interview with  1,274 words and the second interview document contains 2,842 words. 

\begin{figure}[ht!]
\begin{center}
                    \colorbox{lime}{%
    \parbox{\dimexpr\linewidth-2\fboxsep}
        {
        \textbf{Keywords}: family, fun things, friends, strong little community, 
fun games, baseball, 
taxes, more jobs,
Baldwin Reynolds house, park, history, houses
 \\
         \textbf{Summary:} I love all the fun things you can do with friends and family and I feel it's like a strong little community, we all work together and have fun .
What matters to me most is my friends here. 
My favorite memory about living in Meadville is probably playing baseball and doing all these fun games at baseball.
Some stuff to make living here easier would be cutting taxes, creating more jobs , and making it so we're more of a strong  economy.
All the history we have is important , because our history dates clear back to the 1700s . And there 's just so many things you can do like the Baldwin Reynolds House, and even Diamond Park is history.
My one wish for Meadville would be to make all the houses look nice.
        }
    }		
    
    \vspace{1mm}
    
				\colorbox{orange}{%
    \parbox{\dimexpr\linewidth-2\fboxsep}
        {
         \textbf{Keywords:}  community, larger town, 
great education system, crawford county fair,
family-friendly town, great family atmosphere,
economy,  jobs, newer buildings,
good honest living, historical places \\
        \textbf{Summary:} What I love most about the city of Meadville is that it has all of the attractions and items of a much larger town, but it has a very small-town, family-friendly oriented community .
What matters most to me is having a great education system for my children. 
My favorite memory would probably have to be the Crawford County Fair. 
Having a great family atmosphere and a family-friendly town was important to us as we raised our family. Our families grew up here in this area and we 're happy to be around them, and stay in a great community .
Seeing a good, solid base in the economy would be something that would make us stay here.
So between a strong education system and also jobs that people can have that people can make a good honest living off of will keep people here, and keep the town thriving .
What draws us here and keeps us here is all the items of a large town but in a much smaller community oriented town.
 I would probably go with newer buildings versus all the effort and time and money that it would take to restore a lot of these historical places.
        }
    }
\end{center}
\caption{Example output produced by our system for two interview documents.}
\label{fig:output}
\end{figure}

\vspace{-0.1in}	
\subsection{Searching Functionality}
\vspace{-0.02in}	
The summary and keyword information produced by our system was very well received by My Meadville.  In our discussions with them we have estimated savings of over 300 hours of human labor when reading condensed summaries and keywords instead of the complete interviews during their work of compiling value statements and locating supporting data. During one of our feedback discussions, My Meadville administrators requested a possibility of implementing the searching functionality to locate specific information in the summarized output quickly. This was especially important because of a number of new research and community projects in the city that could benefit from the data.   

\begin{figure*}
\begin{center}
	\includegraphics[scale=0.45]{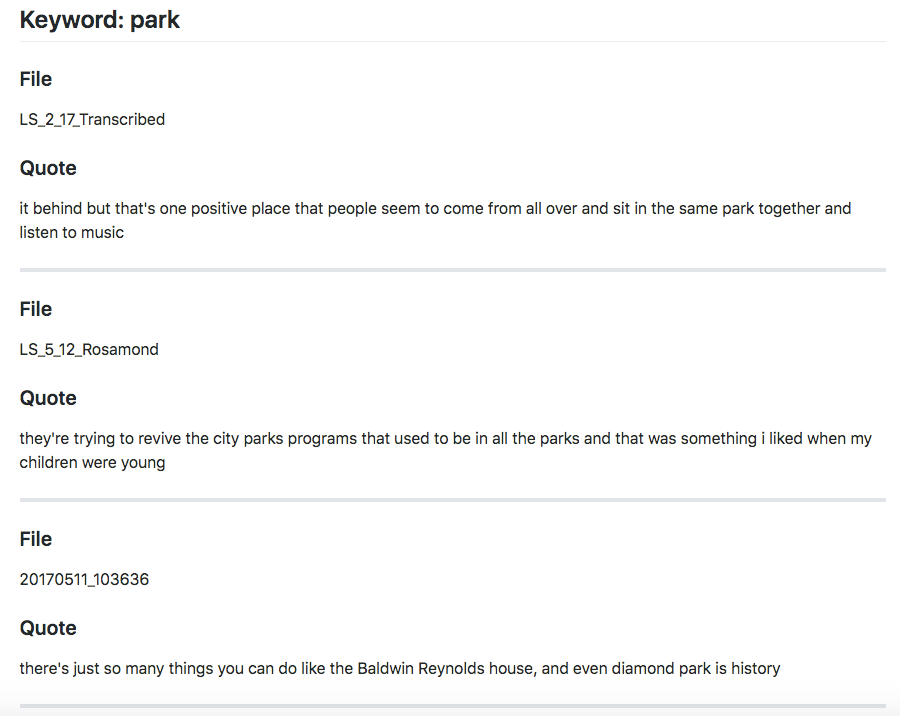}
	\includegraphics[scale=0.45]{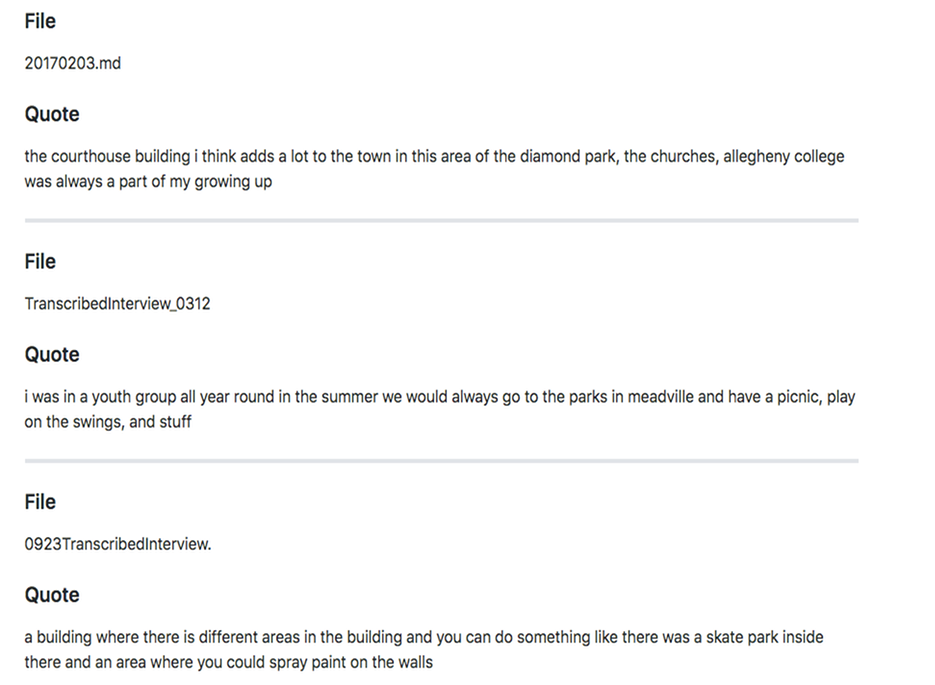}
\end{center}
\caption{An example of the searching output.}
\label{fig:search}
\end{figure*}

We built a supplemental searching tool to accompany our text extraction system that allows the user to search for specific keywords. This tool goes through each output (summary and keywords) produced by our system and tries to match the keyword specified by the user with one of the keywords identified by our system. If there is a match, the keyword, the interview document name, and the text surrounding the keyword is reported in a Markdown file. After the output of all documents has been checked, the Markdown file is uploaded to the GitHub repository, which can be viewed by the relevant parties. An example of an automatically generated and uploaded search result file is shown in Figure \ref{fig:search}, where partial results for the ``park'' keyword are shown. 
Overall, the searching functionality added to our system was proven to be very useful and it allowed the city and the local community to utilize the data for specific projects as needed.

\vspace{-0.1in}	
\section{Conclusion and Future Directions}
\vspace{-0.02in}	
In this paper we describe an intelligent text summarization system that also identifies important keywords and allows to search results for use-specified keywords. We also present an application of this system to the interview data collected by My Meadville initiative, which was supported by the city of Meadville and the City Council and hosted by the Redevelopment Authority of the City of Meadville. Our experimental results were run on over 700 transcribed interviews and the output containing an extracted summary and keywords for each interview were shared with My Meadville committee, which they used in their work of developing community value statements, followed by action plan items. The community work on talking to the Meadville residents continues and more documents are added as the new interviews are transcribed. The searching functionality of our system has been utilized in several community projects, including the study on the enhancement of the city's summer parks program and the feasibility study on the creation of a Community Museum of Science, Industry, and Culture. 

We  encountered a few challenges in our work of AI system application in the public sphere. First of all, we came into the project after the data was collected and hence could not provide input into the transcription format of the interviews and the importance of consistency across transcriptions. Since the interviews were conducted and transcribed by dozens of different volunteers, the structure and the format of the interviews and transcriptions varied greatly. We spent a number of hours manually parsing through interview documents and editing documents that did not identify the interviewee responses easily. Secondly, we found that deploying the system for the use by My Meadville committee was challenging. My Meadville administrators and most of its volunteers had no computing training and were uncomfortable setting up and running programs. Therefore, they relied on us to gather the summary and keywords results. We did successfully train My Meadville administrators in using the searching functionality  as that implementation did not rely on many libraries. We also successfully trained them to use GitHub, where we shared the source code, output files, etc. through private repositories. 

In the future, we would like to extend our system to a container-based set up, where all dependencies will be included for the user in a container and they will not need to download all the dependencies  before using our system. We would also like to create a web-based tool for our searching functionality instead of having an executable file that the users have to click on. Finally, we will continue to coordinate with the City of Meadville in different uses of our system and its further development in order to accomplish their goals of public service enhancements in Meadville.

\vspace{-0.1in}	
\bibliographystyle{aaai} 
\bibliography{bibliography}
\end{document}